\documentclass[runningheads]{llncs}
\usepackage[T1]{fontenc}
\usepackage{graphicx}
\usepackage{amsmath,bbold}
\usepackage{booktabs}
\usepackage{caption}
\usepackage{subcaption}

\usepackage{hyperref}

\usepackage{color}

\begin{document}
\title{$\mu$YOLO: Towards Single-Shot Object Detection on Microcontrollers}
\author{Mark Deutel\inst{1}\orcidID{0000-0001-8932-5212} \and
Christopher Mutschler\inst{2}\orcidID{0000-0001-8108-0230} \and
Jürgen Teich\inst{1}\orcidID{0000-0001-6285-5862}}
\authorrunning{M. Deutel et al.}
\institute{Friedrich-Alexander-Universität Erlangen-Nürnberg, Germany\\ 
\email{\{mark.deutel,juergen.teich\}@fau.de} \and
Fraunhofer Institute for Integrated Circuits, Fraunhofer IIS, Germany\\
\email{christopher.mutschler@iis.fraunhofer.de}}
\maketitle              % typeset the header of the contribution
\begin{abstract}
This work-in-progress paper presents results on the feasibility of single-shot object detection on microcontrollers using YOLO. Single-shot object detectors like YOLO are widely used, however due to their complexity mainly on larger GPU-based platforms. We present $\mu$YOLO, which can be used on Cortex-M based microcontrollers, such as the OpenMV H7 R2, achieving about 3.5 FPS when classifying 128x128 RGB images while using less than 800 KB Flash and less than 350 KB RAM. Furthermore, we share experimental results for three different object detection tasks, analyzing the accuracy of $\mu$YOLO on them.

\end{abstract}
\section{Introduction}

Object detection in computer vision describes the task of first finding, i.e., regressing, a bounding box, and then classifying objects in scenes. Early approaches to this problem use a sliding window to look at evenly spaced locations of the image \cite{felzenszwalb2010objdetect} or, like regional CNNs (R-CNNs) \cite{girshick2014rcnn}, solve the two problems separately by first using some region proposal algorithm and then classifying the generated set of proposals by re-purposing a standard CNN. Nowadays, these approaches are mostly dominated by single-shot detectors (SSDs), with the most prominent one being YOLO \cite{redmon2016yolo}. They are capable of solving the detection task extremely efficient using only a single CNN, and thus do not require a separate region proposal stage. However, even though SSDs are efficient and therefore offer huge potential for energy-efficient real-time object detection, their deployment on edge devices has so far been achieved mainly for larger embedded platforms such as the Jetson Xavier \cite{wong2019YOLONano,yuchen2020mobilenet}, while the feasibility of object detection on more resource-constrained microcontrollers is still an open topic.

We present $\mu$YOLO, optimized for use on Cortex-M based microcontrollers, and show first results from work in progress. %While the use of YOLO on powerful embedded platforms is well explored \cite{wong2019YOLONano}, the feasibility of object detection on more resource-constrained, Cortex-M based systems is still an open topic.
With our work, we aim to evaluate the feasibility and limitations of object detection on microcontrollers, despite severe resource constraints (typically less than 1-2 MB of Flash and less than 1 MB of RAM). To meet these constraints, we lower the input image resolution to $128 \times 128$ pixels, significantly reduce the number of trainable parameters of the backbone network, and reduce the grid size and bounding box predictors per cell compared to the original YOLO architecture. In the following we present a solution deploying $\mu$YOLO on a Cortex-M7 based OpenMV H7 R2 microcontroller, while achieving a framerate of 3.5 frames per second at 480 MHz while requiring less than 800 KB of Flash and less than 350 KB of RAM.
%
%The idea of YOLO and other single-shot detectors is to solve object detection tasks using a single neural network and by looking at given images only once, i.e., in a single forward pass. To achieve this, networks like YOLO use a CNN backbone followed by an object detection head. The detection head typically consists of linear layers and makes predictions in a $S \times S$ grid-based manner. For each grid cell, the network proposes a class prediction vector and a fixed set of bounding boxes $B$, each defined by a confidence score $C$, a center $x, y$ relative to the upper left corner of its cell, and a size $w, h$ relative to the image size. Overall, this results in an output of the form $S \times S \times N + B * 5$ for each image.
%
%To train YOLO, an additive loss is calculated from three components: (1) the mean squared error (MSE) between the predicted centers and dimensions of all bounding boxes responsible for a ground truth box and the actual centers and dimensions of the ground truth boxes, (2) the MSE between the class prediction vectors of all cells containing the center of a ground truth box and the actual class of the ground truth boxes, and (3) the MSE between the predicted confidence scores of predicted bounding boxes and the intersection over union (IoU) with the ground truth boxes, or zero if a predicted box is not responsible for any ground truth box.

The remainder of this paper is structured as follows: First, in Section~\ref{sec:impl}, we discuss the architecture of $\mu$YOLO. Second, in Section~\ref{sec:eval}, we present results from ongoing work on three object detection tasks, two derived from subsets of the COCO dataset \cite{lin2014microsoft} and one using a self-recorded dataset. Furthermore, based on the results obtained so far, we analyze the performance of $\mu$YOLO, provide an error analysis and discuss the limitations of our approach. Section~\ref{sec:conclusion} concludes.
\section{$\mu$YOLO}
\label{sec:impl}

The idea of YOLO is to solve object detection tasks using a single DNN and by looking at images only once, i.e. in a single forward pass. To achieve this, YOLO consists of a CNN backbone followed by a detection head, which typically contains linear layers that make predictions in a $S \times S$ grid-based manner. For each grid cell, the detection head proposes a class prediction vector of size $N$ and a set of bounding boxes $B$, each defined by a confidence score, a center point relative to the upper left corner of its cell, and a size relative to the image size.

The backbone CNN of $\mu$YOLO consists of a single convolution, followed by seven depth-wise separable convolutions \cite{chollet2017xception}, see Table~\ref{tab:arch} for a detailed description of the architecture. We choose depth-wise separable convolutions to minimize trainable parameters. $\mu$YOLO's classification head has two linear layers and has an output similar to the original YOLO paper \cite{redmon2016yolo} with a shape of $S \times S \times N + B * 5$. Our experiments so far have shown that, given the resource constraints of our target platform and the low spatial resolution of the input images ($128 \time 128$), $B=2$ and $S=5$ is the best compromise. We use ReLU as the activation, but use batch normalization instead of a dropout layer before the last linear layer as described in the original YOLO paper.

We pre-trained our backbone model on the Caltech-256 dataset \cite{griffincaltech2562007}, which is a 256 class image classification problem. We achieved a top-1 accuracy of $~38\%$ and a top-5 accuracy of $~61\%$ with $\mu$YOLO's backbone, which while below other large-scale classification models, is still reasonable given the small capacity of our backbone and the large number of classes in Caltech-256.

To further reduce the size of $\mu$YOLO, we apply network pruning to all convolutional and linear layers during training using an iterative gradual pruning schedule \cite{zhu2017prune} with a $L1$ norm heuristic, as well as 8-bit quantization afterwards. We did not use quantization-aware training in order to minimize the overhead required for training, as we realized early on that for our application it did not provide an improvement over post-training quantization. To automate these steps and to convert the resulting model to deployable C-code, we apply the compression and deployment pipeline described in \cite{deutel2022deployment}.

\begin{figure}[t]
    \centering
    \begin{minipage}[t]{.44\textwidth}
        \centering
        \captionsetup{type=table}
        \captionof{table}{Architecture of $\mu$YOLO. Each row describes a layer, we omit activation functions (ReLU) and batch normalization layers for brevity. The tuples describe the convolutional layer ("C") and the depthwise seperable convolutional layers ("D") in the form of [channels, filters, kernel size, stride, padding] while they describe linear layers ("L") in the form of [num. input, num. output].}
        \label{tab:arch}
        \begin{tabular}{c}
        \toprule
        \textit{Input:} $3 \times 128 \times 128$ \\
        \midrule
        {C: [3, 64, 4, 2, 0]}  \\
        \midrule
        MaxPool: {[2]} \\
        \midrule
        {D: [64, 128, 3, 1, 0]}  \\
        {D: [128, 128, 3, 1, 1]}  \\ 
        {D: [128, 128, 3, 1, 0]}  \\
        \midrule
        MaxPool: {[2]} \\
        \midrule
        {D: [128, 128, 3, 1, 1]}  \\
        {D: [128, 64, 3, 1, 0]}  \\ 
        {D: [64, 64, 3, 1, 1]}  \\
        {D: [64, 64, 3, 1, 0]}  \\
        \midrule
        MaxPool: {[2]} \\
        \midrule
        {L: [1024, 1024]} \\
        {L: [1024, $S \times S \times N + B * 5$]} \\
        \midrule
        \textit{Output:} $S \times S \times N + B * 5$ \\
        \bottomrule
        \end{tabular}
    \end{minipage}
    \hfill
    \begin{minipage}[t]{.54\textwidth}
        \centering
        \begin{minipage}[t]{\textwidth}
            \centering
            \captionsetup{type=table}
            \captionof{table}{Performance in frames per second (FPS) and memory consumption of $\mu$YOLO after pruning and 8-bit quantization on the OpenMV H7 R2 microcontroller for the three detection tasks considered.}
            \label{tab:deploy}
            \begin{tabular}{l c c c}
            \toprule
            {} & \textbf{Fridge} & \textbf{Vehicles} & \textbf{Humans}\\
            \midrule
            \textbf{Flash} & 752 KB & 771 KB & 706 KB\\
            \textbf{RAM} & 324 KB & 324 KB & 324 KB\\
            \textbf{FPS} & 3.46 & 3.45 & 3.49\\
            \bottomrule
            \end{tabular}
        \end{minipage}\vspace{4mm}
        \begin{minipage}[t]{\textwidth}
            \centering
            \captionsetup{type=table}
            \captionof{table}{Performance in frames per second (FPS) and memory consumption of $\mu$YOLO for the vehicles task at different input image resolutions. Where deployment on the target system was not feasible due to memory constraints, no FPS value is provided.}
            \label{tab:deploy_size}
            \begin{tabular}{l c c c}
            \toprule
            {} & \textbf{Flash} & \textbf{RAM} & \textbf{FPS}\\
            \midrule
            \textbf{88x88 px} & 133 KB & 152 KB & 8.40\\
            \textbf{128x128 px} & 771 KB & 324 KB & 3.45\\
            \textbf{256x256 px} & 774 KB & 1.01 MB & -\\
            \textbf{448x448 px} & 774 KB & 3.97 MB & -\\
            \bottomrule
            \end{tabular}
        \end{minipage}
    \end{minipage}
\end{figure}

\section{Evaluation}
\label{sec:eval}

We evaluated $\mu$YOLO on three detection tasks: (1) human detection, (2) vehicle detection (truck, bus, car, bicycle, motorcycle), both derived from the Microsoft COCO dataset \cite{lin2014microsoft}, and (3) toy groceries detection in a mini-fridge (water, milk, chocolate milk, orange juice) using a dataset we recorded ourselves. For all tasks, we used an input resolution of $3 \times 128 \times 128$ and applied affine transformations as well as brightness, saturation, and hue adjustments to all training samples during training to prevent overfitting. We trained all models for 400 epochs using stochastic gradient descent (SGD) with learning rate of $0.001$, momentum of $0.9$, and weight decay of $0.005$. During the last 100 epochs of training, we applied iterative pruning every 20 epochs. Furthermore, due to the low resolution of the input, we discarded ground truth boxes that became smaller than $12 \times 12$ pixels.

\begin{figure}[t]
    \centering
    \begin{minipage}[t]{.49\textwidth}
        \captionsetup{type=figure}
        \includegraphics[width=\textwidth]{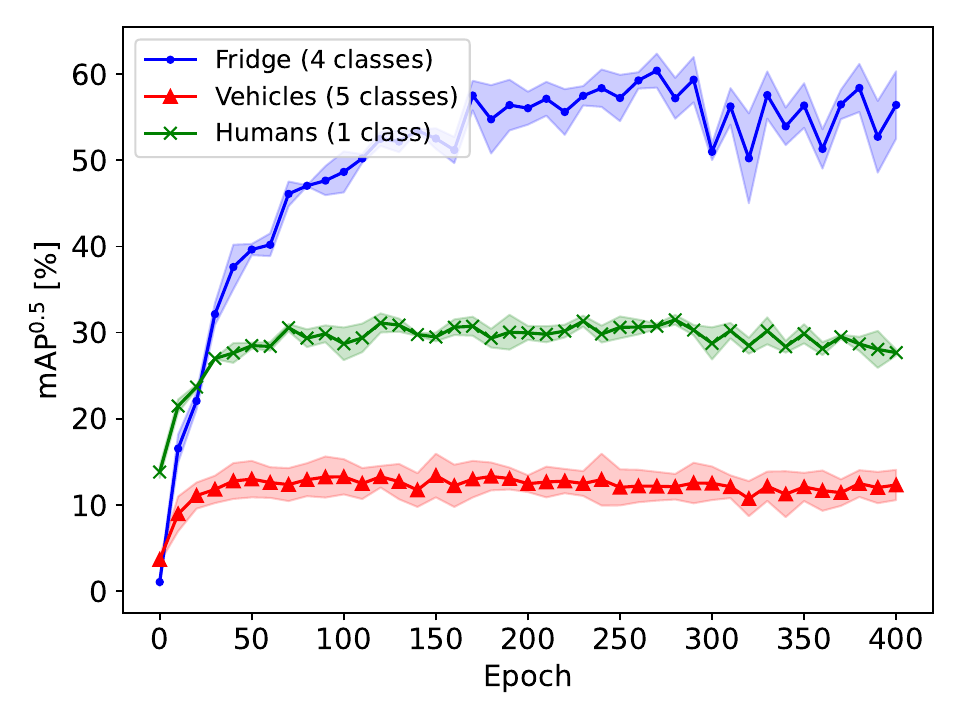}
        \captionof{figure}{Average $\mathrm{mAP}^{0.5}$ achieved on the validation data of the considered object detection tasks over the course of 400 training epochs. 3 seeds for each task.}
        \label{fig:map_results}
    \end{minipage}\hfill
    \begin{minipage}[t]{.49\textwidth}
        \captionsetup{type=figure}
        \includegraphics[width=\textwidth]{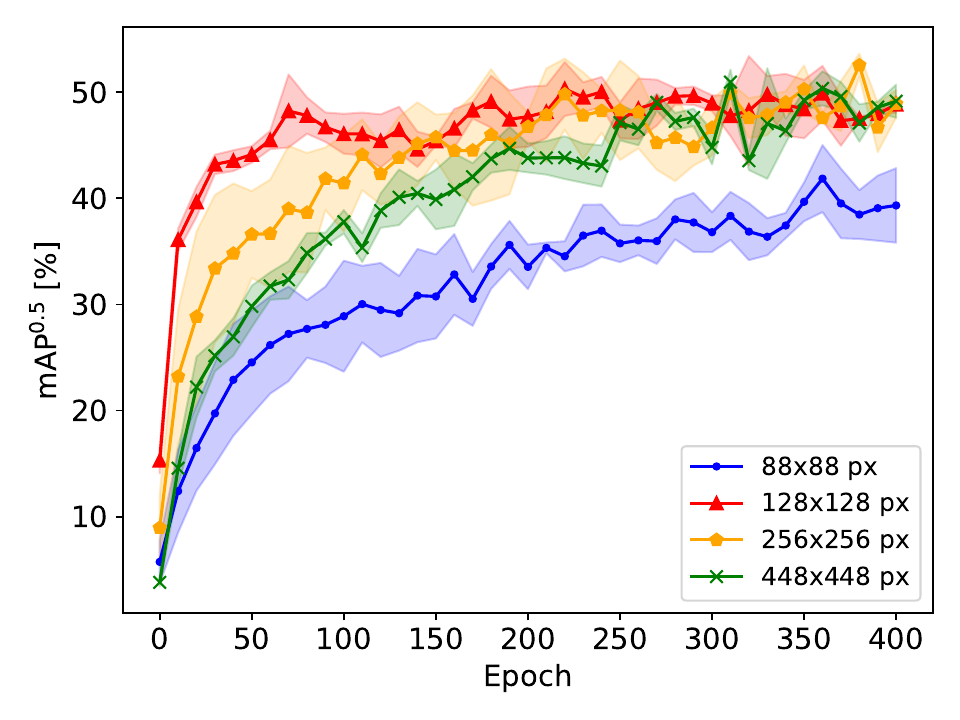}
        \captionof{figure}{Average $\mathrm{mAP}^{0.5}$ achieved on the validation data of the vehicles task given different input image sizes and assuming a maximum of 3 bounding boxes per image. 3 seeds for each task.}
        \label{fig:map_results_size}
    \end{minipage}
\end{figure}

\subsection{Experimental Results}

We present the mean average precision with an intersection over union (IoU) threshold of 0.5 ($\mathrm{mAP}^{0.5}$) over the course of training for all three detection tasks, see Fig.~\ref{fig:map_results}. We repeated the training for all tasks three times with different seeds. Overall, we observed that $\mu$YOLO achieved a significantly higher $\mathrm{mAP}^{0.5}$ score on the refrigerator detection task (56.4\% on average) than on the other two tasks (27.7\% on average on the human detection task and 12.3\% on average on the vehicles detection task). We observed a slight degradation of $\mathrm{mAP}^{0.5}$ especially for the fridge task (blue curve) during the last 100 epochs of training due to pruning, while we did not notice any significant degradation after quantization. 

We argue that the strong contrast in performance observed in the three different tasks shown in Fig.~\ref{fig:map_results} is mainly the result of the different scene depth and complexity present in the datasets, combined with the low input resolution, which makes it difficult to detect and distinguish particularly small objects in the background.
To test this hypothesis, we trained $\mu$YOLO on a simplified version of the vehicle task and at different input image resolutions, considering as ground truth only a maximum of three largest bounding boxes per image, see Fig.~\ref{fig:map_results_size}. To handle larger input image resolutions, we increased the kernel size in the first convolution of $\mu$YOLO to 7 and the padding of the following two depthwise-separable convolutions to 2. For smaller image resolutions, we decreased the number of neurons in the linear layers of the detection head.

Looking at Fig.~\ref{fig:map_results_size}, $\mu$YOLO was able to achieve a significantly higher $\mathrm{mAP}^{0.5}$ on the simplified vehicles detection task than on the unrestricted version (red curve in Fig.~\ref{fig:map_results}) for all tested input image resolutions. Interestingly, while choosing an extremely small input resolution had a negative impact on the achieved precision, see the blue curve compared to the other three curves, increasing the image resolution did not affect the observed $\mathrm{mAP}^{0.5}$, which always converged to 50\%, see the red, orange, and green curves. Comparing these three curves, we were even able to observe that larger image resolutions slowed the convergence rate at the beginning of training. Therefore, we conclude that while the overall complexity of the scene as well as the size of the objects relative to the image size have a major impact on the performance of $\mu$YOLO, increased image resolutions do not. When also considering memory consumption, see Table.~\ref{tab:deploy_size}, an input resolution of 128x128 proved to be the most feasible trade-off.

\begin{figure}[t]
    \centering
    \begin{subfigure}[t]{\textwidth}
        \centering
        \includegraphics[width=\textwidth]{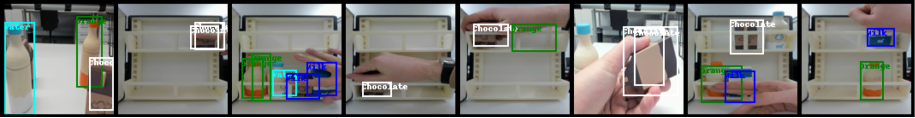}
        \caption{Fridge (light blue - water, blue - milk, white - chocolate milk, green - orange juice)}
        \label{fig:qual_fridge}
    \end{subfigure}
    \begin{subfigure}[t]{\textwidth}
        \centering
        \includegraphics[width=\textwidth]{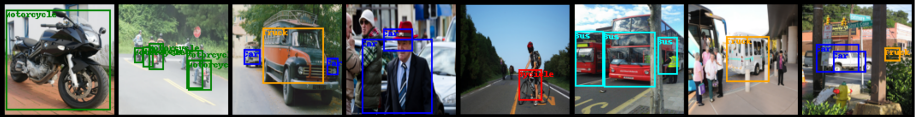}
        \caption{Vehicles (light blue - bus, blue - car, orange - truck, red - bicycle, green - motorcycle)}
        \label{fig:qual_streets}
    \end{subfigure}
    \begin{subfigure}[t]{\textwidth}
        \centering
        \includegraphics[width=\textwidth]{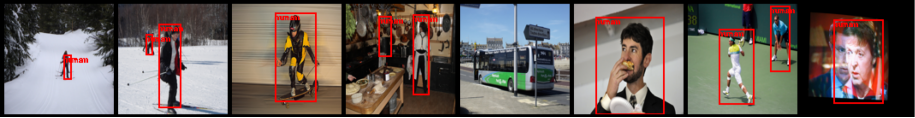}
        \caption{Humans (red - human)}
        \label{fig:qual_humans}
    \end{subfigure}
    \caption{Samples from the validation sets of the three considered detection tasks, adding to them all bounding boxes that achieved a confidence greater than  50\%.}
    \label{fig:qual_results}
\end{figure}

In addition, to give a better intuition of how well $\mu$YOLO works, we also provide qualitative results on a subset of samples from the validation datasets of the three unrestricted detection tasks, see Fig.~\ref{fig:qual_results}. We noticed that when looking at the qualitative results, the large difference in $\mathrm{mAP}^{0.5}$ that can be observed between the detection tasks in Fig.~\ref{fig:map_results} is not so obvious.
In general, for all the tasks, most of the bounding boxes predicted by $\mu$YOLO are relatively accurate, with only a few misclassifications that we noticed. The most prominent source of error we observed, is that $\mu$YOLO fails to detect present objects, especially when they are partially occluded or further in the background.

\begin{figure}[t]
     \centering
     \begin{subfigure}[t]{0.32\textwidth}
         \centering
         \includegraphics[width=\textwidth]{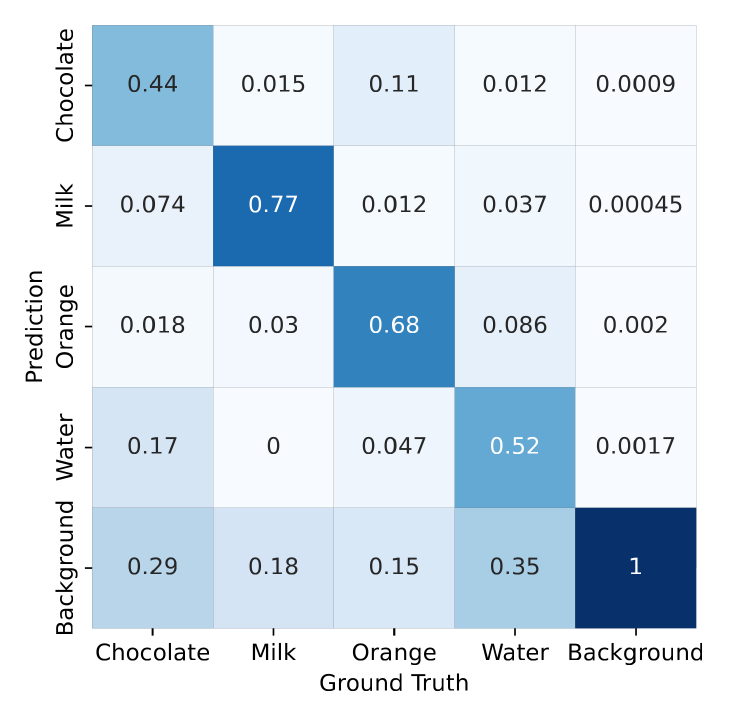}
         \caption{Fridge}
         \label{fig:confusion_fridge}
     \end{subfigure}
     \hfill
     \begin{subfigure}[t]{0.32\textwidth}
         \centering
         \includegraphics[width=\textwidth]{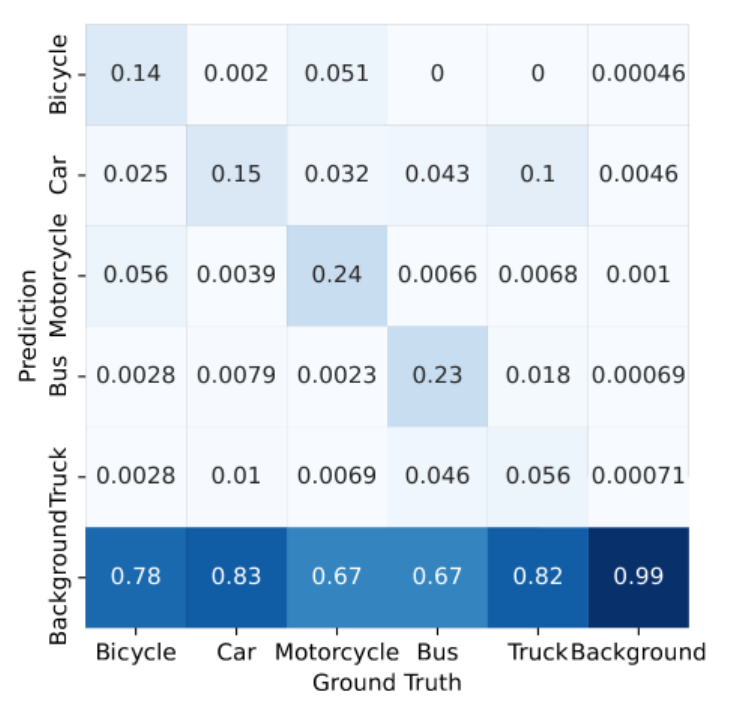}
         \caption{Vehicles}
         \label{fig:confusion_street}
     \end{subfigure}
     \hfill
     \begin{subfigure}[t]{0.32\textwidth}
         \centering
         \includegraphics[width=\textwidth]{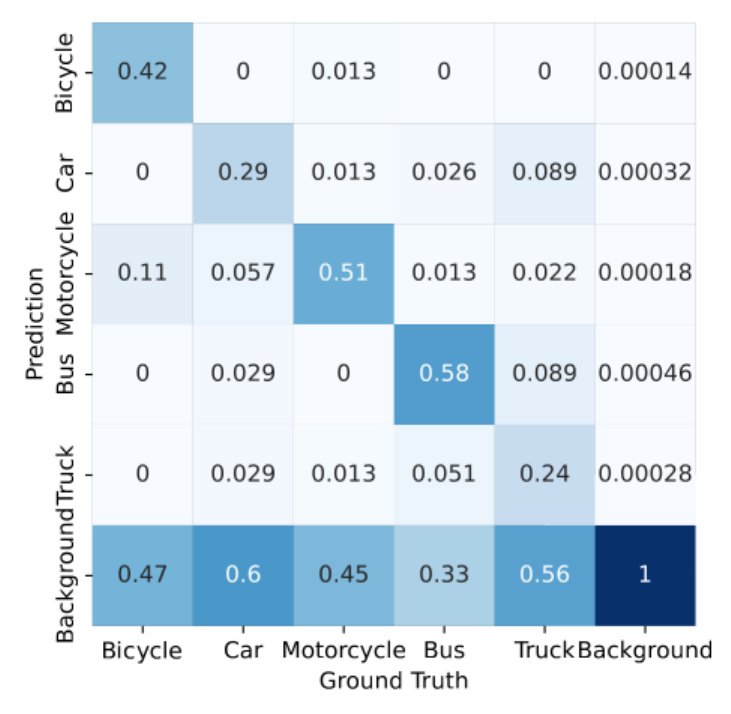}
         \caption{Vehicles (Max. 3 boxes)}
         \label{fig:confusion_street_reduced}
     \end{subfigure}
     \caption{Normalized confusion matrices of predicted bounding boxes versus ground truth. The diagonals are correct predictions, while the upper and lower triangular matrices are errors. A bounding box is correct if its class prediction is correct and both its confidence and IoU with the ground truth are greater than 50\%.}
     \label{fig:confusion}
\end{figure}

\subsection{Error Analysis}

To deeper investigate made observations, we analyzed the detection errors of $\mu$YOLO by computing the confusion matrices over the validation datasets for the fridge and vehicles tasks, see Fig.~\ref{fig:confusion}. Each of the confusion matrices represent the predicted bounding boxes by class, i.e. rows, compared to the actual ground truth bounding boxes provided by the validation set, i.e. columns. Thereby, a predicted bounding box is "not background" if its confidence score is higher than 50\%, and it is a correct prediction if it overlaps its corresponding ground truth with an IoU higher than 50\%. This means that, the diagonals of the matrices contain the correctly predicted bounding boxes, while the triangular matrices above and below contain errors. More precisely, the last row of each matrix denotes false negatives, the last columns denotes false positives, and all other fields denote misclassifications of correctly detected bounding boxes. 

Comparing Fig.\ref{fig:confusion_fridge} with Fig.~\ref{fig:confusion_street} and Fig.~\ref{fig:confusion_street_reduced}, the most common observable errors of $\mu$YOLO on the validation sets were false negatives, i.e. $\mu$YOLO "missing" present objects. Other errors, such as misclassifying correctly detected objects or mistakenly detecting an object where none exists, i.e. false positives, were much less common in comparison. However, we found that these effects were much more pronounced in the unrestricted vehicles detection task, see Fig.~\ref{fig:confusion_street}, than in the restricted vehicles and unrestricted fridge detection tasks, see Fig.~\ref{fig:confusion_fridge} and Fig.~\ref{fig:confusion_street_reduced}. Based on these observations, we argue that the high number of false negatives we found in Fig.~\ref{fig:confusion_street} is mainly due to the low input resolution of $\mu$YOLO combined with the high complexity of multi-layered scenes with object overlap and partial occlusion, as they are predominantly present in the COCO dataset. In comparison, the fridge task or the simplified vehicles task are less complex, which as a result seems to allow $\mu$YOLO to achieve higher $\mathrm{mAP}^{0.5}$ scores, mainly due to false negatives being less common.

\subsection{Deployment}

To deploy $\mu$YOLO on the OpenMV H7 microcontroller we used the pipeline described in \cite{deutel2022deployment}, see Table~\ref{tab:deploy} and Table~\ref{tab:deploy_size} for a detailed breakdown of memory consumption and performance. The difference in Flash consumption and slight change in performance that can be observed in Table~\ref{tab:deploy} is due to the size of the output feature map being dependent on the number of classes, thereby changing the number of neurons in the output layer. Furthermore, in Table~\ref{tab:deploy_size} it can be seen that changing the size of input images processed by $\mu$YOLO has a big impact on RAM consumption and performance. In particular, the large increase in RAM consumption makes the use of $\mu$YOLO with larger input image resolutions on the OpenMV H7 microcontroller quickly infeasible.

Besides changing the input image resolution, the trade-off between FPS and $\mathrm{mAP}^{0.5}$ can also be controlled via pruning. In initial experiments and by using aggressive pruning schedules, we were able to achieve up to 8 FPS, but at the cost of a significantly degraded $\mathrm{mAP}^{0.5}$. Since 3.5 FPS might be too low for some detection applications, we consider further exploration of the trade-off between model performance and precision as future work.
\section{Conclusion}
\label{sec:conclusion}

We presented $\mu$YOLO, a novel single-shot object detection model based on YOLO that can be deployed on microcontroller platforms without hardware acceleration with a memory footprint less than 800 Kb of Flash and less than 350 Kb of RAM. We presented work in progress results on three different object detection tasks and provide an in-depth analysis of the accuracy $\mu$YOLO can achieve on these tasks. We also shared results from deploying $\mu$YOLO on a Cortex-M7 based "OpenMV H7 R2" microcontroller, where we achieved a performance of about 3.5 frames per second for all three tasks. 

%\subsubsection{Acknowledgements} Please place your acknowledgments at
%he end of the paper, preceded by an unnumbered run-in heading (i.e.
%3rd-level heading).

\bibliographystyle{splncs04}
\bibliography{literature}
\end{document}